\documentclass[conference]{IEEEtran}
\IEEEoverridecommandlockouts
\usepackage{cite}
\usepackage{amsmath,amssymb,amsfonts}
\usepackage{algorithmic}
\usepackage{graphicx}
\usepackage{textcomp}
\usepackage{xcolor}
\usepackage{enumitem}
\usepackage{balance}

\newif\ifhighlight
\highlightfalse % Uncomment to disable highlighting

\newcommand{\highlight}[1]{\ifhighlight\textcolor{magenta}{#1}\else#1\fi}

\def\BibTeX{{\rm B\kern-.05em{\sc i\kern-.025em b}\kern-.08em
    T\kern-.1667em\lower.7ex\hbox{E}\kern-.125emX}}
\begin{document}

\title{The \textit{Beatbots}: A \highlight{Musician-Informed} Multi-Robot Percussion Quartet
% {\footnotesize \textsuperscript{*}Note: Sub-titles are not captured in Xplore and
% should not be used}
% \thanks{Funded by university grants.}
}

% \author{
% Anonymous authors
% }

\author{
\IEEEauthorblockN{Isabella Pu}
\IEEEauthorblockA{\textit{MIT Media Lab} \\
Cambridge, USA \\
ipu@media.mit.edu}
\and
\IEEEauthorblockN{Jeff Snyder}
\IEEEauthorblockA{\textit{Princeton University} \\
Princeton, USA \\
josnyder@princeton.edu}
\and
\IEEEauthorblockN{Naomi Ehrich Leonard}
\IEEEauthorblockA{\textit{Princeton University} \\
Princeton, USA \\
naomi@princeton.edu}
}

\maketitle
\IEEEpeerreviewmaketitle

\begin{abstract}

Artistic creation is often seen as a uniquely human endeavor, yet robots bring distinct advantages to music-making, such as precise tempo control, unpredictable rhythmic complexities, and the ability to coordinate \highlight{intricate} human and robot \highlight{performances}. While many robotic music systems aim to mimic human \highlight{musicianship}, our work emphasizes the unique strengths of robots, \highlight{resulting in a novel multi-robot performance instrument called the \textit{Beatbots}, capable of producing} music \highlight{that} is challenging for \highlight{humans} to replicate \highlight{using current methods}. \highlight{The \textit{Beatbots} were designed using an ``informed prototyping'' process, incorporating feedback from three musicians throughout development.} We evaluated the \textit{Beatbots} through a live public performance, surveying participants ($N=28$) to understand how they perceived and interacted with the robotic performance. \highlight{Results show that participants valued the playfulness of the experience, the aesthetics of the robot system, and the unconventional robot-generated music. Expert musicians and non-expert roboticists demonstrated especially positive mindset shifts during the performance, although participants across all demographics had favorable responses.} We propose design principles to guide the development of future robotic music systems \highlight{and identify key robotic music affordances that our musician consultants considered particularly important for robotic music performance.}

\end{abstract}

\begin{IEEEkeywords}
robotic music, multi-robot systems, informed prototyping
\end{IEEEkeywords}

\section{Introduction}
Creating art is often seen as a uniquely human \highlight{venture}, tied to emotion, creativity, and expression \cite{hoffmann2021emotions}. Yet, the intersection of robotics and art is expanding \cite{gemeinboeck2013creative,jeon2017robotic,thorn2020human}, challenging traditional ideas on art and sparking excitement and skepticism about robots' artistic role. \highlight{In this work, we focus on robotic music, an area where} robots offer distinct advantages, such as the creation of unpredictable rhythms \cite{weinberg2006toward, weinberg2007robotic} and precise tempo control \cite{crick2006synchronization}, expanding possibilities in musical performance. 

\highlight{While} many current robotic music systems are designed to replicate human playing, often being trained on human musicians to sound as ``human'' as possible \cite{savery2021shimon,hoffman2011interactive,kapur2007integrating,kusuda2008toyota}, \highlight{our} approach emphasizes \textit{robotic} \highlight{capabilities}. \highlight{We explore using robots as an alternative means of music creation, focusing on performing music that would be challenging for humans to replicate alone.} By integrating human-robot co-creation principles \cite{candy2002integrating,gomez2021robot} and utilizing robotic abilities like the addition of randomness \cite{bruun2020human} and coordination between multiple robots within a system \cite{albin2012musical}, we showcase how robots can contribute to music \highlight{in novel ways}, while still including meaningful human interaction \cite{rowe2004machine} \highlight{and, importantly, centering human musicians' values \cite{poirson2007integration}}. 

For this work, we chose percussion music because it aligns well with known robotic strengths. Opting for a multi-robot system also allowed us to explore more complex rhythmic coordination between musicians, as seen in several well-known contemporary percussion compositions \cite{clapping,glass}. We specifically chose the \highlight{recognized} percussion quartet form \cite{ice2012percussion}.

Our robotic percussion quartet, called the \textit{Beatbots}, \highlight{represents a novel artistic system for producing and performing robotic music.} To ensure alignment with human musicianship values, we collaborated with three musicians \highlight{through informed prototyping \cite{camburn2017design} in the design of our robotic music system.} \highlight{Our work} was driven by the following research questions:

\begin{enumerate}[label=R\arabic*)]
    \item \textbf{Through \highlight{informed prototyping} with musicians, can we develop a \highlight{novel} robotic system that utilizes robots' unique strengths \highlight{in percussion music performance?}}
    \item \textbf{How do experts and non-experts in music and robotics perceive robotic percussion music performance?}
    \item \textbf{What do participants value in a multi-robot percussion music experience? How does interaction with robotic percussion performers affect that experience?}
\end{enumerate}

We also present a public demonstration and evaluation of the \textit{Beatbots} system, providing insights into how different demographics---experts and non-experts in percussion music and robotics---perceive the \highlight{performance}. \highlight{Our work also introduces five design principles for robotic percussion systems based on insights from the demonstration and discusses musical robot affordances that emerged from our musician-informed design process. We aim to offer reusable strategies and insights for future robot music performance development.}

\section{Related Work}

\subsection{Algorithmic \& Generative Percussion Music}

Percussion music originated from instinct, with early percussionists relying on innate musicality. Over time, techniques were codified and passed down over generations \cite{hartenberger2016cambridge, benzon1993stages}, with rhythmic repetition and rule-based phrasing forming a foundation for intricate patterns \cite{adler1999mathematics}. This enabled the creation of new, complex patterns derived from traditional phrases, much like certain procedural approaches to music composition \cite{langston1989six}, which range from medieval to contemporary works \cite{mcalpine1999making}. Additionally, contemporary percussion composers sometimes move away from strict rules, introducing elements of randomness to add further musical complexity \cite{tucker2017emergence,popoff2011indeterminate}.

Steve Reich is known for repetitive rhythms and phasing \cite{reich}, where phrases are played at different tempos to create desynchronized textures, as seen in  \textit{Drumming} and \textit{Clapping Music} \cite{clapping, hartenberger2016performance}. Similarly, Philip Glass uses overlapping rhythms to build complex layers from simple patterns, as in his \textit{String Quartet 6} \cite{isac2020repetitive, glass}. John Cage, in works like his \textit{Composed Improvisations}, pioneered rule-driven compositions \cite{feisst2009john} that use randomness, allowing performers to improvise within structured frameworks---such as having musicians roll dice to decide their next pattern \cite{cage}.

Our work takes inspiration from contemporary percussion compositions by incorporating phasing, rhythmic layering, and randomness, implemented by our robotic music system. These techniques, pioneered by composers like Reich, Glass, and Cage, are particularly well-suited for our approach emphasizing unique robot strengths, as robots can easily execute random or complex patterns requiring difficult coordination, like phasing and overlapping rhythms, which would require extensive training for human musicians to perform accurately. 

\highlight{\subsection{Robot Musicianship}}

As robots become \highlight{increasingly prevalent} \cite{cone2019robots,jimeno2019fewer}, their \highlight{applications} in creative fields \highlight{extend} beyond automation \cite{gemeinboeck2013creative,jeon2017robotic,thorn2020human,savery2021shimon}. \highlight{Musical robots have been developed for various instruments \cite{bretan2016survey, kapur2005history}, ranging from string \cite{kusuda2008toyota, jorda2002afasia}, to wind \cite{dannenberg2011mcblare,solis2006waseda}, to percussion \cite{hoffman2010shimon,weinberg2006robot}.} 

Recent \highlight{developments in} robotic music \highlight{tend to emphasize \textit{anthropomorphism}, with systems mimicking human appearance} or trained on human \highlight{movements} \cite{savery2021shimon,hoffman2010shimon,kapur2007integrating}. \highlight{Examples include the Waseda Flutist Robot \cite{solis2006waseda}, Toyota's violin-playing robot \cite{kusuda2008toyota}, and others \cite{uchiyama2023development,zhang2011musical, wu2010towards}. Robot systems also leverage physical presence and visual cues---important elements to live performance \cite{schutz2008seeing}---which digital musicians cannot \cite{weinberg2007robotic, pessanha2021virtual}.}

%, which use anthropomorphism as a vehicle toward more familiar and ``human'' music performances

\highlight{While robotic music systems are often viewed as novel instruments \cite{solis2011musical}, musicians emphasize that human control---either through programming or real-time interaction---is essential for classification as a new instrument \cite{weinberg2007robotic}. Systems like Shimon \cite{hoffman2010shimon, hoffman2011interactive} or GuitarBot, in performance with violinist Mari Kimura \cite{auslander2009lucille}, demonstrate real-time interaction through their ability to synchronize and adapt to human musicians.}

\highlight{Robots also possess capabilities that transcend human limitations, such as increased accuracy \cite{zhuo2021human}, improved ability to follow} instructions while introducing \highlight{unpredictable} variations \cite{weinberg2007robotic, bruun2020human}, \highlight{and the ability to perform feats of speed or scale that would require complex coordination of multiple humans \cite{weinberg2006toward, bretan2016survey}. For example, Haile, a robot that plays a Native American drum, exceeds human speed \cite{weinberg2006toward}, while other systems utilize three-dimensional space to achieve orchestral effects \cite{bretan2016survey, flo2015doppelganger}. However, these systems typically employ instrument actuation methods that mimic human movements, still aiming to produce music that sounds as human-like as possible.}

% Our work diverges from prior approaches by explicitly leveraging robots' unique capabilities rather than emulating human musicianship. Through the Beatbots system, we explore new artistic possibilities enabled by algorithmic percussion music and novel actuation methods. While this approach may produce unconventional musical output, we believe it will advance the development of distinctly robotic music systems. Throughout our design process, we prioritized input from human musicians [51] to ensure our system maintains artistic integrity while pushing the boundaries of traditional musicianship.

\highlight{Our work diverges from prior approaches by explicitly leveraging robots' unique capabilities rather than emulating humans. Through the \textit{Beatbots}, we explore new artistic possibilities enabled by our algorithmic percussion music and novel actuation method. While this approach may produce unconventional musical output, we believe it will lead to new insights toward distinctly robotic music systems and musical culture of the future \cite{rowe2004machine}. Given this divergence, we prioritized input from human musicians \cite{vear2024jess} throughout our design process to ensure our system maintains artistic integrity while pushing the boundaries of traditional musicianship.}

\section{Design Method}
\highlight{This section outlines the design process of the \textit{Beatbots}, a four-robot system which leverages robot strengths by playing algorithmic percussion music using whole-body kinetic movement. Robot behaviors are inspired by leader-follower rules, and they can be interacted with through moving the robots and arenas or controlling robot behavior via a keyboard interface.}

\subsection{\highlight{Informed Prototyping}}
\highlight{When developing the \textit{Beatbots}, our primary objective was to center the values of its ideal users: musicians \cite{kujala2003user,zamenopoulos2018co}. Unlike technology-focused research that typically prioritizes system accuracy and efficiency \cite{turchet2018smart}, designing novel musical systems that prioritize \textit{user needs} requires a deep understanding of what matters most to users \cite{friedman1996value}, often diverging significantly from designers' initial goals \cite{turchet2018smart}. As robotic systems become increasingly sophisticated, incorporating user values into the design process becomes even more critically important \cite{sellen2009reflecting}.}

\highlight{We therefore involved musicians as key consultants in} defining the robotic system \highlight{through \textit{informed prototyping} \cite{camburn2017design}}. This approach enabled a nuanced integration of human creativity \highlight{and musician values} with machine potential, ensuring the system was grounded in users' artistic and ethical values \cite{halloran2009value}.

\highlight{We sought feedback from three musicians during the design process: one classically trained percussionist (M1), one guitarist, composer, and instrument designer (M2), and one pianist formally trained in both classical and jazz piano (M3). All three have some level of experience with percussion playing, are practicing live performers, and compose music. Additionally, M2 earned a doctoral degree in a music domain and M3 earned a Bachelor's degree from a music conservatory.}

\highlight{We gathered feedback from our musician consultants at various stages of the design process, with detailed results presented in the following sub-sections. Specifically, we sought feedback for robot choice, programmatic music composition, instrument selection, arena design, autonomous robot behaviors, and methods of human control over the robots. We collected this feedback by presenting the current prototype to the musicians in-person as a live performance, allowing them to share their thoughts verbally without any guiding questions. This approach helped identify what aspects were most important to the musicians without external influence. After hearing initial insights, researchers asked follow-up questions to clarify and ensure a thorough understanding, enabling us to effectively and accurately integrate their feedback into the system.}

\subsection{Robot Choice}

An early design choice \highlight{by the research team} was to make the robots identical in appearance and sensors, creating a cohesive look \highlight{and enabling future scalability.} \highlight{Additionally,} rather than using conventional stationary actuators, like a robot arm, \highlight{our musicians advocated for leveraging the robots' ability for dynamic, whole-body movement, and instead} use kinetic motion to strike instruments\highlight{---a distinctly non-anthropomorphic approach}. \highlight{They specifically favored} \textit{Sphero}\footnote{https://sphero.com} robots over more traditional vehicle-like designs, \highlight{citing their novel, futuristic aesthetic.} Using rolling balls to hit drums along the arena's sides was \highlight{also} likely to be more \highlight{visually} engaging, especially for non-technical viewers. We chose the \textit{Sphero BOLT}s\footnote{https://sphero.com/collections/all/products/sphero-bolt} for their built-in sensors, including a gyroscope and motor encoder, and programmable display lights, which added visual interest and interactivity. We hypothesized that these features would enhance the experience for non-technical audiences.

\subsection{Percussion Music}\label{percuss-design}

\begin{figure}[htbp]
\centering
\includegraphics[width=0.485\textwidth]{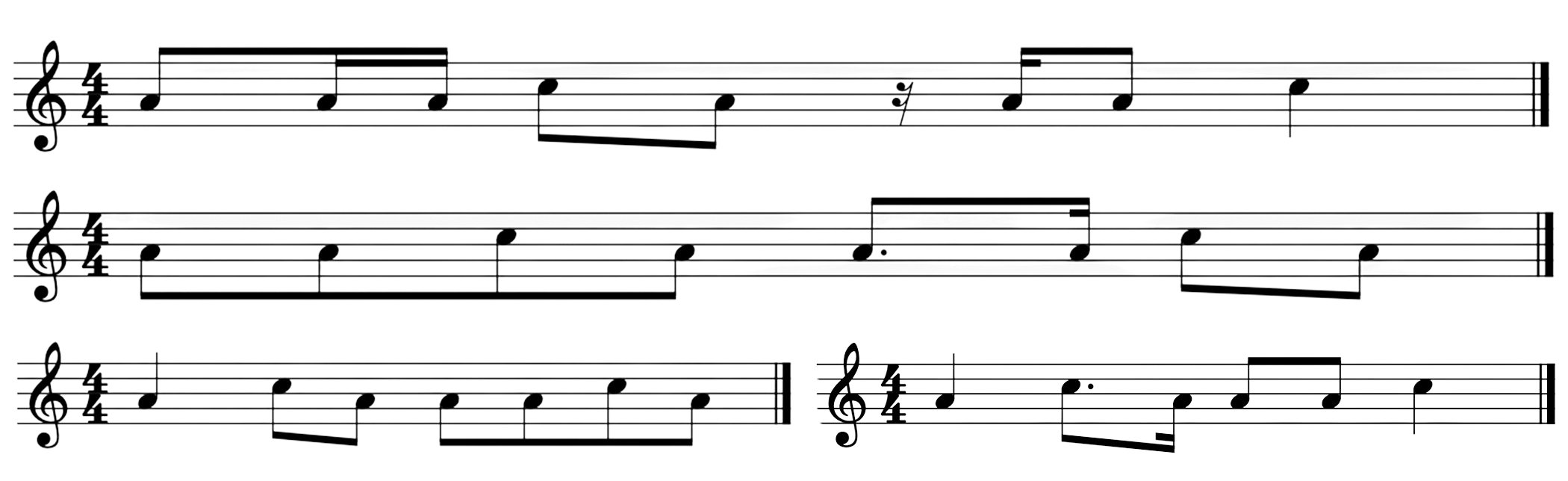}
\caption{Example of four basic patterns provided by \highlight{M1, a percussionist}}
\label{fig:patterns}
\end{figure}

Based on \highlight{feedback from} our musician \highlight{consultants}, we \highlight{prioritized the robots' capability} to \highlight{play} authentic percussion rhythms. Similar to Cage's \textit{Composed Improvisations} \cite{cage} \highlight{and other works illustrating the divisible nature of music \cite{benford2012supporting}}, the system’s music was built from pre-written, rule-based percussive patterns selected randomly. These patterns were informed by interviews with our \highlight{musician consultants, particularly M1}, through which we gathered twenty-five percussive rhythms \highlight{which the \textit{Beatbots} could combine to} create a rich and complex musical performance (example rhythms in Fig. \ref{fig:patterns}). 

With the help of our \highlight{musician consultants, particularly M1,} we categorized these patterns. Uneven groupings, such as \textit{long, short, short, shortest, long, shortest, short}, were \highlight{categorized as} ``uneven patterns'', while even groupings like \textit{long, long} \highlight{were} ``even patterns''. We also differentiated between quicker and slower patterns---``slower'' featured mostly longer note values, while ``quicker'' contained more shorter note values.

Another key concept gleaned from our \highlight{musicians} was the distinction between \textit{single} and \textit{double} strokes. A single stroke occurs when one downward motion creates a single sound, while a double stroke uses the momentum of the first hit to bounce and produce a second sound. A single stroke is heard as a longer sound, whereas a double stroke is heard more as two shorter sounds. The percussion patterns \highlight{our musicians} suggested included both single and double strokes, requiring additional considerations in robot behavior.

Instrument choice was another area where we sought input from \highlight{musicians}, piloting several instruments with them before making a final decision. We tested pitched (chimes, kalimba) and unpitched instruments (frame drums, shakers, cymbals). While pitched instruments added melody, their shape often caused the robots to veer off-course after impact, and incorrect pitches were far more sonically disruptive than mistimed unpitched sounds. Ultimately, we chose unpitched instruments, finding frame drums were the best-sounding option and could also easily function as walls for the four-sided robot arena.

We pilot-tested several drum sizes and found the ten-inch drums produced the best sound when struck by the robots. Frame drums also offered the advantage of producing two distinct tones: a deeper ``bass'' tone when hit in the center and a sharper ``slap'' tone when struck near the edge. By positioning the drums accordingly, both sounds could be played despite the robots' small size. The original plan was to make all four arena walls out of ten-inch frame drums. However, the robots often accidentally hit drums that were not their direct target while moving toward their ``goal'' drum, making it difficult for audiences to distinguish between what we dubbed ``purposeful'' and ``accidental'' sounds.

\highlight{Our musicians noted that percussionists in traditional quartets often play multiple instruments, inspiring our solution: } we introduced ten-inch tambourines \highlight{to the arena, positioning identical instruments across from each other}. The tambourine’s metal jingles added a distinct sound, differentiating ``accidental'' hits on tambourines from ``purposeful'' ones on frame drums. \highlight{This} also \highlight{enabled} robots to switch instruments mid-performance, adding a dynamic element to the composition.

\subsection{``Arena'' Design}

The robots' enclosure, or ``arena,'' played a key role in shaping the rhythms the robots could produce with kinematic motion and whether single or double strokes could be played.

\begin{figure}[htbp]
\centering
\includegraphics[width=0.335\textwidth]{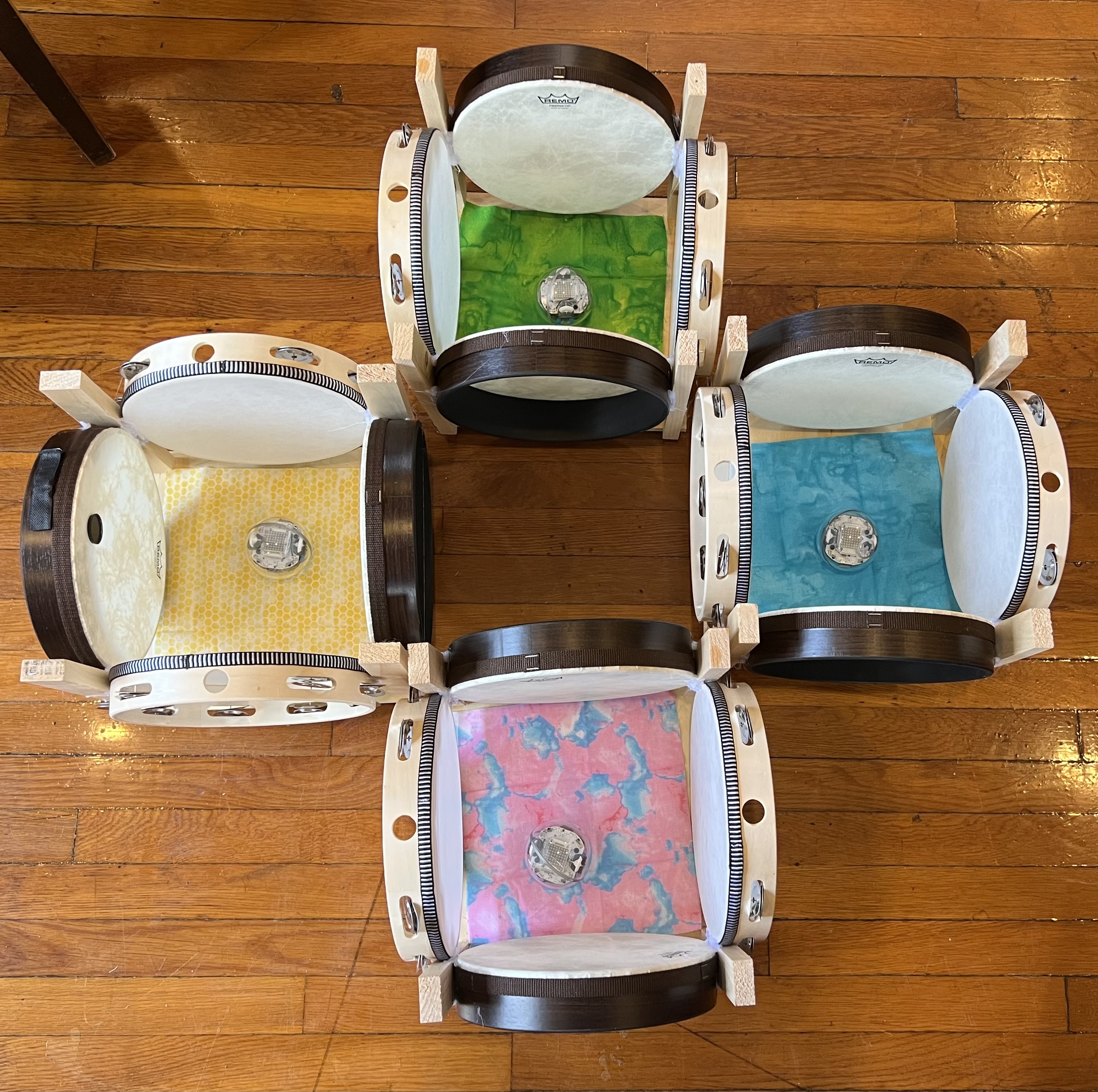}
\caption{Example layout of the four robot arenas.}
\label{fig:four-robots}
\end{figure}

From the start, instruments were positioned along arena walls to optimize acoustics, as drums placed perpendicular to the robots produced the loudest sound. Each robot started the performance in the center of its own arena, with instrument placement and arena size refined through pilot testing. After testing sizes up to two feet by two feet, we settled on ten inch by ten inch arenas, with each wall made near-entirely of a frame drum or tambourine, supported by small wooden bars.

This compact design allowed for flexible arrangements, with \highlight{easily movable arenas} to accommodate different spaces. For instance, arenas could be arranged as shown in Fig. \ref{fig:four-robots} for a clear view of all robots at once or positioned to surround the audience for an immersive experience. \highlight{Our musicians suggested using distinct colors to help audiences identify individual ``performers''. We therefore assigned unique colors to each robot-arena pair, helping viewers recognize each distinct unit in the quartet despite a uniform physical appearance.}

\begin{figure*}[htbp]
\centerline{\includegraphics[width=0.85\textwidth]{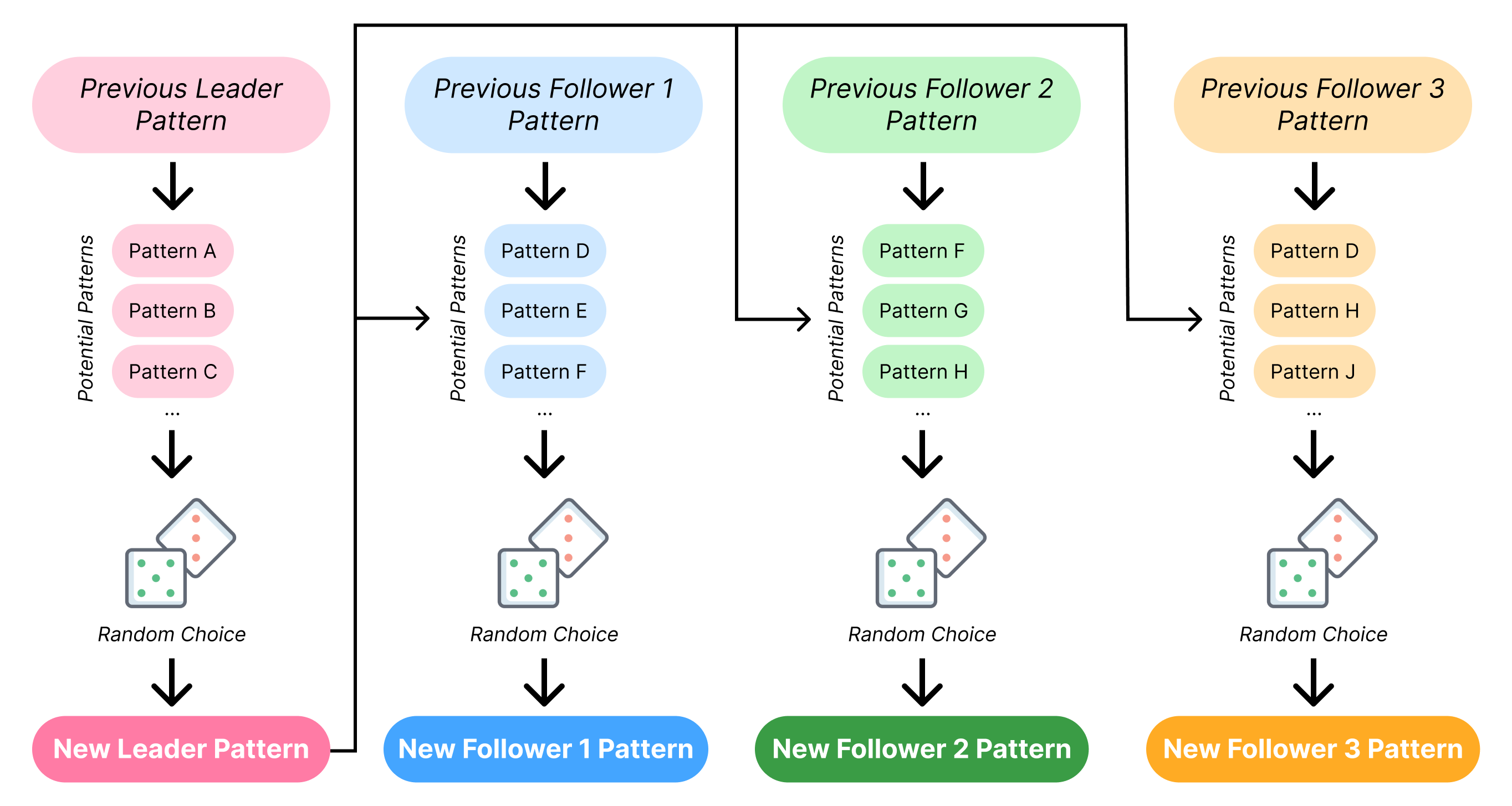}}
\caption{Diagram of how robots in the quartet choose their next percussion pattern. Each robot took into account its previous pattern, and all follower robots took into account the leader robot's current pattern.}
\label{fig:system-fig}
\end{figure*}

\subsection{Robot Behaviors}\label{robot-design}

We implemented \highlight{two} distinct roles within our quartet, designating one robot as ``leader'' and the others as ``followers.'' This structure is common in group music settings, such as with lead singers, conductors, and concertmasters \cite{benzon1993stages}, and it guided the selection of rhythmic patterns. Each robot, at every step of the performance, had a list of potential patterns that it randomly selected its next pattern from. This list was based on each robots' previous instruction and---if it was a follower---the leader's current instruction. This approach mirrors contemporary techniques, like in Cage's \textit{Composed Improvisations} \cite{cage}, where the score is actually a set of instructions. These leader-follower rules were developed in collaboration with \highlight{musician consultants}. Fig. \ref{fig:system-fig} contains a diagram depicting how the robots chose their next pattern.

To ensure the robots played basic rhythms as accurately as possible, we implemented a global ``metronome-clock'' synchronized to real-world time. This clock maintained a tempo of 60 beats per minute (BPM), guiding the robots when playing their selected pattern. We found that, despite the slow 60 BPM tempo, the robots struggled to play rhythms accurately due to their imprecise movement control. Even with periodic re-synchronization using the global ``metronome-clock,'' the robots quickly fell out of sync. For example, even small deviations in timing when aiming to spend 0.25 seconds on a note became quite noticeable within one or two patterns.

While these instabilities initially sounded jarring, they created interesting patterns when played in a group---reminiscent of the phasing seen in Reich's \textit{Clapping Music} \cite{reich,clapping} and the sound layering of Glass' \textit{String Quartet 6} \cite{glass}. \highlight{When presented with the robot music, our musicians} encouraged us to embrace these sonic complexities as the timing inconsistencies introduced a unique \textit{desynchronous} element---one that would be difficult for humans to consistently replicate, which was our design goal. Unlike humans, robots could add or subtract truly random intervals between notes, even doing so while playing in a group, without being influenced by other performers.

Through experimentation, we also found that the robots' spherical shape helped produce both single and double strokes. A single stroke occurred when the robot maintained forward motion after hitting the drum, whereas for a double stroke, cutting motor power immediately after impact caused the robot to bounce back then forward again for a second, softer hit. While unintended double strokes sometimes occurred (for instance, if power was cut slightly too early), this unexpected behavior further added sonic complexity, which our \highlight{musician consultants} saw as a positive contribution to the composition.

Finally, to address situations where the robots failed to detect collisions after hitting the walls, we implemented a fail-safe. If no collision was detected after four seconds---the length of one pattern---the robot would automatically turn around and begin their next pattern. This sometimes led to unintended silences, but \highlight{our musicians} felt that having moments of rest added a natural musical element.

\subsection{\highlight{User Interface}}\label{user-interface}

Our musician \highlight{consultants} emphasized the need for incorporating human interaction into the robotic system’s behavior. \highlight{This would not only enhance audience engagement but also help musicians feel actively involved in making the music. They suggested that interacting with the system felt like programming a synthesizer or creating a DJ set---while robots physically made the music, musicians saw the system as a new method of music creation under their artistic control.}

The initial design allowed users to press buttons at different points in the performance to control the robots' movements, preserving the generative composition while offering some user control. Pilot testing revealed that using a digital piano keyboard for this interaction made \highlight{participants} feel \highlight{more musically engaged}, transforming the interaction from a distant performance to a more personal musical experience.

Implemented human \highlight{interactions} fell into two categories: changing the robots' \textit{lights} and their \textit{movements}. By default, the robots' display lights shifted to a new hue of their assigned color every second. However, when different piano keys were pressed, the robots synchronously changed their lights to various new colors, allowing users to control an important visual element of the performance. Pilot testing showed visual control was particularly influential for non-musician users.

Pressing different piano keys also allowed users to control various aspects of robot movement. Users could make the robots spin in place, move in circles around their arena, switch primary instruments---changing between frame drum and tambourine by turning 90 degrees---or return to the center of the arena. Users could also stop and restart the robots' default behavior at any time. A sequence of stopping, re-centering, and restarting helped manage desynchronization and any other minor differences in the global ``metronome-clock,'' allowing users to adjust the performance’s musical complexity.

\section{User Study}

\begin{figure}[thbp]
\centering
\includegraphics[width=0.4\textwidth]{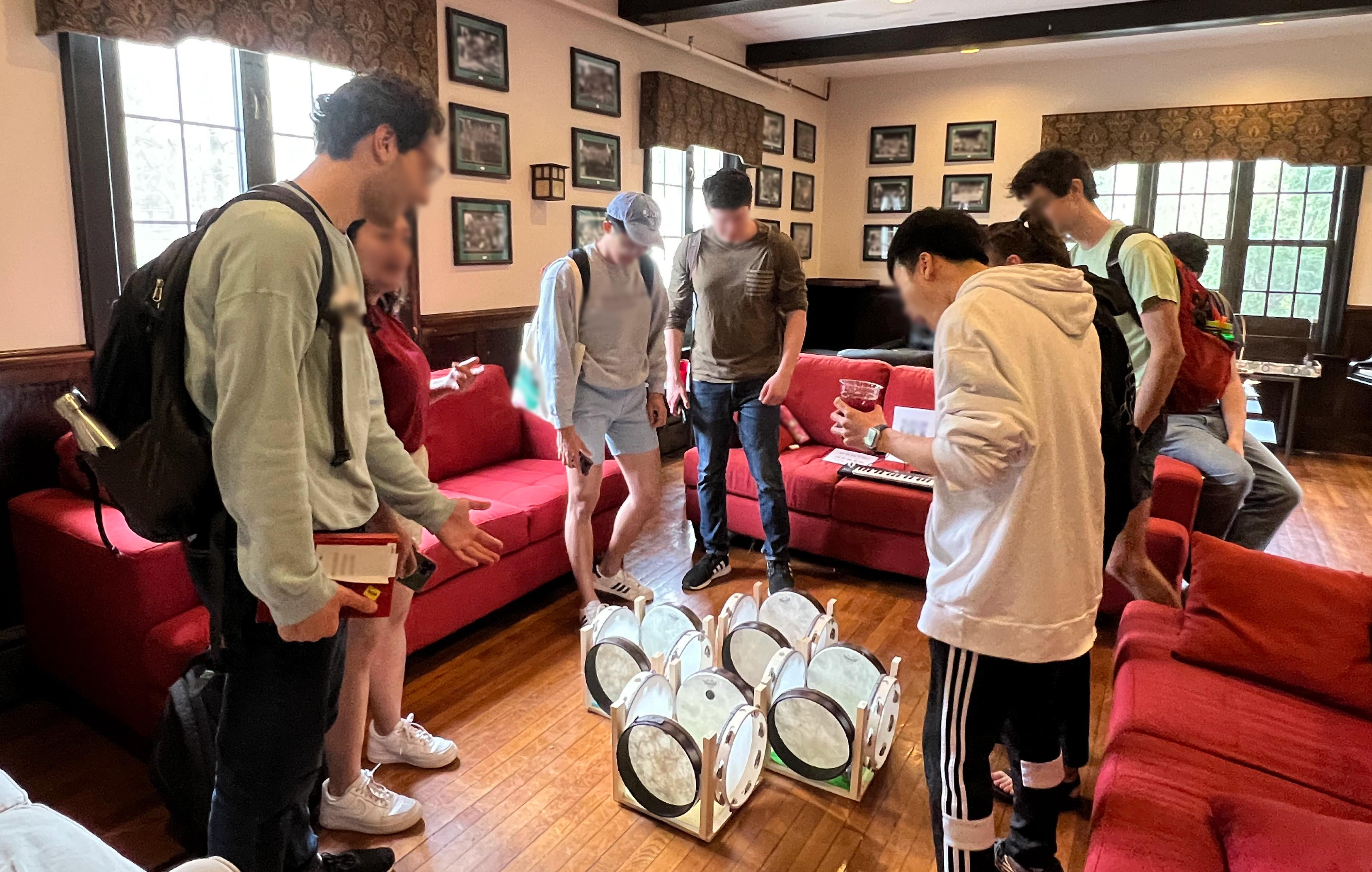}
\caption{A scene from the live robotic percussion quartet performance.}
\label{fig:photo}
\end{figure}

To investigate different populations' perceptions of the robotic quartet, we conducted a user study inviting university students and community members to view and \highlight{interact with} the robots at a local arts space, as seen in Fig. \ref{fig:photo}.

Study activities and surveys were reviewed by a university Institutional Review Board.

\subsection{Participants}

$40$ community members attended with $28$ completing our survey. The performance was advertised via university mailing lists and bulletin boards. All $28$ survey respondents were students of the university where the performance was located. Each participant was asked to rate their familiarity with percussion music and robots on a five-point Likert scale, with average familiarity with percussion music being $2.14$ ($SD=1.18$) and familiarity with robots being $1.89$ ($SD=1.13$). 

\subsection{Study Design}
Participants were invited to \highlight{first} watch the \textit{Beatbots} perform \highlight{live}, with the option to \highlight{control parts of the performance} afterwards. Almost all participants ($N=24$) chose to try \highlight{controlling parts of the robot musical performance}.

\subsubsection{Robotic Performance}
The \textit{Beatbots} played a generative percussion composition which participants would first watch for one to three minutes without interacting with the robots.

\subsubsection{\highlight{Non-Performance} Interaction}
Participants were invited to pick up the robots, play with the robots \highlight{outside the arenas}, and move the robot arenas around the room as they saw fit.

\subsubsection{Performance \highlight{Interaction}}
Participants were then invited to \highlight{control} the \textit{Beatbots} in \highlight{another} performance. They were encouraged to try all available methods of \highlight{controlling} the performance, as described in section \ref{user-interface}: changing the robots' colors, switching the robots' primary instrument, making the robots spin in place, making the robots travel in circles around the arena, stopping then restarting the robots' movements, and re-centering the robots. These contributions were all mediated by pressing keys on a small piano keyboard.

\subsubsection{User Survey}
At the end of the experience, participants were asked to fill out an optional user survey \highlight{which was disseminated anonymously through a nearby QR code}. The survey asked them to rate their familiarity with percussion music and robots each on a five-point Likert scale. Additionally, they were asked three free-response questions: \textit{``Did you enjoy the experience? What made it positive or negative?''}, \textit{``Do you believe a human quartet could play what you just heard?''}, and \textit{``Did you enjoy the human interaction component? How did it add or detract from the experience?''}

\subsection{Data Collection \& Analysis}

Researchers at the performance took detailed observational notes, and survey responses were collected at the scene.

\begin{table*}[ht!]
\caption{Overview of Themes and Sub-themes Found in Participant Survey Responses}
\label{tab:survey-codes}
\centering
\resizebox{\linewidth}{!}{
\begin{tabular}{l|l|l}
\hline
\textbf{Themes}& \textbf{Sub-themes}& \textbf{Definitions}                                                          \\
\hline
                                    \highlight{Playful Robot Engagement}& \highlight{Robot Control Enjoyment}& \highlight{Enjoyed controlling the robots}\\
 & Robot Interaction Enjoyment&Enjoyed interacting with robots \highlight{in non-control ways}\\
                                    & Robot Movement Enjoyment& Enjoyed how the robots moved around the arena\\
                                    \hline
 \highlight{Robot Aesthetic Appeal}& \highlight{Visual Engagement}&\highlight{Engaged by the robot visuals (e.g., lights or colors)}\\
 & \highlight{Found Robots ``Cute'' \& ``Funny''}&\highlight{Appreciated robots' friendly and humorous behavior}\\
 & Calming Experience&Described the experience as calming or peaceful\\
                                    \hline
 \highlight{Musical Appreciation}& Percussion Music Enjoyment&Enjoyed how the percussion music sounded\\
 & \highlight{Appreciation of Complex} Rhythms&\highlight{Specifically appreciated the \textit{complexity} of rhythms}\\
 \hline
\end{tabular}}
\end{table*}

In our survey analysis, we conducted inductive analysis \cite{guest_applied_2012} of participants' free responses to generate a list of relevant themes. Two researchers independently reviewed responses, inductively identified, \highlight{and finalized} themes. Then, two independent researchers coded survey responses \highlight{with} the themes, and we calculated Cohen’s Kappa to measure inter-rater reliability, achieving a ``substantial'' agreement score of \highlight{$\kappa = 0.76$} (``substantial'' agreement referring to $0.61 < \kappa < 0.80$) \cite{mchugh_interrater_2012}.
% and resolved all disagreements to reach a consensus \cite{daly_design_2012,ostrowski2022mixed}.

\section{Results}

\highlight{Analysis of participant responses revealed several key findings about user experience. We first present themes identified through analysis of survey free responses (see Table \ref{tab:survey-codes}), followed by patterns observed across different levels of expertise.}

\subsection{\highlight{Playful Robot Engagement}} 

\highlight{Both the performance and interactive aspects of the \textit{Beatbots} resulted in an engaging experience characterized by what several participants described as ``playfulness'' in robot interaction, control, and autonomous movement.} P23 remarked, \textit{``I found it to be childlike... I enjoyed the aspect of play.''}

The performance evoked strong positive responses, many of which cited \highlight{robots' rolling movement as a reason for enjoyment.} P6 wrote \textit{``I enjoyed seeing them roll around,''} \highlight{while P16 appreciated }\textit{``that [the robots] roll around in little tambourine cages.''} \highlight{The coordinated movement between robots especially impressed viewers, with} P11 noting they \textit{``liked that each of the robots worked together to create a cohesive piece.''}

Most participants did not immediately \highlight{approach the interactive opportunities}. However, after researchers mentioned the \highlight{possibility for interaction}, nearly all respondents $(N=24)$ tested the human-robot interactions. Of those, \highlight{the vast majority} reported positive experiences, with \highlight{only two expressing reservations, stating that control} made the experience feel \textit{``less magical''} (P24) or \textit{``more predictable''} (P23).

\highlight{The piano keyboard interface for controlling the robots proved particularly engaging.} P2 wrote \textit{``I loved the interactions I had with robot by using the keyboard,''} while P3 remarked that \textit{``it was interesting seeing how the piano was able to control the pattern.''} \highlight{During the interactions,} several participants also remarked that using the keyboard made them feel like fellow musicians \highlight{within the ensemble. P21 expressed the playful nature of controlling the ensemble, writing} \textit{``It made me feel like the director of my own symphony!''}

\highlight{Physical, non-controlling interactions also emerged as a key element of engagement. Many participants valued being able to pick up and move the robots, which seemed to} help them connect with the robots as entities rather than just machines. \highlight{P13 captured this sentiment, saying it was} \textit{``fun to interact with the little robots, really brought joy.''} \highlight{P1 echoed this:} \textit{``It was fun to play around with how the robots acted.''}

\subsection{\highlight{Robot Aesthetic Appeal}}

\highlight{The visual design elements of the \textit{Beatbots} helped craft a distinct aesthetic experience that participants found appealing. The combination of lights, movement, and physical design contributed to both the robots' perceived personality and the overall atmosphere of the performance.}

\highlight{The robots' lights were particularly compelling, featured in several responses. P1 noted how the colored lights contributed to giving the robots personality, observing that they} \textit{``seemed to have some personality when they spinned around and changed color.''} P21 also expressed that they \textit{``enjoyed the colors and rhythms. It gives rhythm and blues a whole new meaning!''}

\highlight{Participants also responded well to the overall charming aesthetic character of the spherical robots. Multiple participants used endearing descriptors,} with P4 calling them \textit{``so cute''} and P8 finding them \textit{``so fun.''} \highlight{This aesthetic encouraged engagement, evidenced by several participant comments, such as} P5 noting they \textit{``like the funny noises and robots.''}

\highlight{The combined aesthetic effects of the robot system created a} calming atmosphere. P14 characterized the experience as \textit{``very zen,''} \highlight{while P25 appreciated it as} \textit{``an engaging sensory experience.''}  \highlight{The careful integration of visuals and other aspects of physical form with the generated music seemed to transform what could have been a purely rhythmic performance into a comprehensive multi-sensory performance.}

\subsection{\highlight{Musical Appreciation}}
\highlight{Participants' engagement with the musical aspects centered around two distinct elements: enjoyment of the percussion music itself and the complexity of the generated rhythms.}

\highlight{Those who praised the music output often described it as unconventional but still appealing.} P14 observed that \textit{``though the sounds sound a bit off tempo, it sounds very calming,''} \highlight{suggesting an appreciation for the atypical rhythms. P27 expressed surprise at the musicality of the system,} noting that \textit{``it was cool that it actually produced a tune.''} \highlight{These responses indicate that participants found value in the robots' musical capabilities, even when---and sometimes because---the output deviated from traditional percussion performance.}

\highlight{The rhythmic complexity also surfaced as an engaging element separate from music enjoyment. When asked whether humans could replicate the performance, participants' responses revealed a fascination with its technical and rhythmic complexity.} P26 noted this, suggesting \textit{``All rhythms are unique and there's probably some 127-th tuplet that would be almost impossible to recreate,''} \highlight{referring to an exceptionally precise musical note value. P22 also pointed to the irregular timing, stating that} \textit{``the rhythms seem a little too erratic/unstable/stop-and-go for a normal human.''} \highlight{Yet rather than viewing these characteristics negatively, participants appreciated them as unique features of robotic musicianship,} comparing the music to ``rave'' or ``experimental'' music \highlight{that moves in unexpected but still engaging ways.}

% Additionally, researchers at the performance observed that the majority of questions posed to the researchers were regarding the subject of reproducibility. Participants asked questions like \textit{``What are robots good at that humans aren't?''} and \textit{``What would a human quartet have to do to sound like this?''}

\subsection{\highlight{Patterns Across Expertise}}

\highlight{A particularly striking finding was the contrast between expert and non-expert responses to the robotic performance. While} nearly all survey respondents ($N=27$) reported a favorable opinion---with one respondent \highlight{reporting} a neutral opinion, \highlight{who had} a previous aversion to percussion music---\highlight{their reasoning varied based on expertise level.}

\highlight{The robot music sharply divided participants along the lines of musical expertise---experts tended to not enjoy the music while non-experts consistently expressed positive reactions. Among our participants,} non-experts ($18$ participants who rated themselves $1$ or $2$ on a five-point Likert scale) provided all positive reviews that specifically mentioned music \highlight{enjoyment} in the free responses, \highlight{indicating a strong enjoyment of the music despite not having the technical background to understand the rhythmic patterns behind the music.}

\highlight{In contrast,} percussion experts ($4$ participants who self-rated as a $4$ or $5$, \highlight{and whom researchers identified as music students}) \highlight{followed a different trajectory.} Researchers who spoke with expert musicians before the performance began noted that expert musicians were far more negative about their hopes for the experience than non-experts. However, by the end of the performance, all four expert musicians who completed the survey indicated positive reactions to the \highlight{complete audio-visual} experience.
\highlight{Though no music experts reported enjoyment of the \textit{music}, they developed appreciation through non-musical aspects. Their music expertise also led them to recognize technical complexities that non-experts missed---for instance, all experts thought the performance would be impossible for humans to replicate exactly, while} all five participants who strongly believed it could be reproduced were non-experts.

\highlight{Robotics expertise also influenced engagement with the system, albeit differently compared to musical expertise. Among our participants,} we had 4 robotics experts (rated $1$ or $2$ on the Likert scale) and 20 non-experts (rated $4$ or $5$ on the Likert scale) based on self-reported expertise.

Robotics experts focused more on the control and coordination aspects, with all experts highlighting their experience controlling the robots in the survey---\highlight{usually expressing a desire for more granular control. This reflects a deeper technical understanding of robots' movement capabilities.} Additionally, researchers at the performance who spoke with expert and non-expert roboticists noted that roboticists actually had higher expectations for what the robots would be able to do, compared to non-experts who were initially more skeptical.

\highlight{Despite this initial skepticism, non-expert roboticists showed more eventual enthusiasm for the robots' movements and physical presence in their responses.} Notably, all participants who explicitly enjoyed the robot movement were non-experts, \highlight{indicating a strong mindset shift regarding robot movements. They also} frequently compared the robotic sounds to familiar musical genres, potentially indicating a focus on the overall experience rather than its technical aspects. \highlight{This focus may also explain why the keyboard interface proved particularly effective at engaging non-experts, as by allowing them to directly control and interact with the robots, the interface provided a hands-on way for non-experts to connect with the performance and appreciate the robots' capabilities, without needing to understand underlying technical complexities.}

\section{Discussion}

\highlight{The complex, unconventional rhythms generated by the \textit{Beatbots} demonstrate the potential for robotic music systems to push artistic boundaries while centering musician values. The positive response to the \textit{Beatbots}' unique robotic rhythms suggests that there is a receptive audience for experimental robotic music. However, the differences in how surveyed experts and non-experts perceived the performance show that, to fully realize the potential of robotic music as an emerging art form, designers must strike a balance between pushing creative boundaries and ensuring that their work resonates with both musicians \textit{and} audience members of all demographics.}

\subsection{\highlight{The Role of Expertise}}

\highlight{Regarding \textbf{R2}, the differences in how music and robotics experts versus non-experts experienced the robotic performance suggest that \textbf{expertise significantly shapes user perceptions}.}

\highlight{These differences highlight the complex role of expertise in shaping audience perceptions of robot music. Initial skepticism of musical experts, stemming from attunement to precision and conventional rhythms, contrasts with the openness of non-experts to unconventional music. Similarly, the expectation of robotics experts regarding technical aspects, like granularity of control, differed from non-experts' initial skepticism and eventual captivation with the ``magical'' robot movements.}

\highlight{Our findings reveal how different aspects of the performance shifted different expert mindsets: hands-on interaction transformed robotics non-experts from skeptical to enthusiastic, while multi-sensory elements won over music experts despite their continued skepticism about the musical output. This demonstrates that expertise does not determine audience responses in a uniform way, with various performance elements creating unique pathways to appreciation. Ultimately, the fact that both experts and non-experts found value in the \textit{Beatbots} performance, even if through different routes, highlights the inclusive potential of robotic musicianship. This suggests that by centering the values of both experts and non-experts, designers can create experiences that push artistic boundaries while still fostering broad engagement and appreciation.}

\subsection{\highlight{Musician-Informed Robot Affordances}}

\highlight{During our design process, our musicians provided valuable insight into what types of capabilities are most important to musicians for inclusion in future robot music systems.} \highlight{To address \textbf{R1}, we provide a list of the musical robot affordances in our system inspired by our musicians' feedback.}

\subsubsection{\highlight{\textbf{Authentic rhythmic patterns}}} \highlight{The actual rhythmic or melodic content used by robotic music systems should be derived from real musical practices, even if methods of playing the music are dissimilar to human methods. This was the most important affordance to our musician consultants.}

\subsubsection{\highlight{\textbf{Music-informed decision-making architecture}}} \highlight{The system's decision-making process should reflect how musicians make choices in musical settings---and if there are multiple robots in the system, they should coordinate in ways inspired by music ensembles, such as leader-follower \cite{benzon1993stages}.}

\subsubsection{\highlight{\textbf{Movement-music integration}}} \highlight{Robot movements should be grounded in musical principles, even when exploring novel capabilities. Our musicians valued how robots' unique movements could parallel traditional musical techniques---like stopping motions mirroring musical rests and the robots' rolling movements enabling single and double strokes.}

\subsubsection{\highlight{\textbf{Visual identity system}}} \highlight{Robot music systems should provide clear visual cues just as traditional music performances do \cite{weinberg2007robotic, pessanha2021virtual}. Visual cues in robotic music systems can enhance audience engagement, particularly for listeners who may be less interested in algorithmic music \cite{schutz2008seeing}.}

\subsubsection{\highlight{\textbf{Musical control interfaces}}} \highlight{Systems should provide meaningful human interaction and control that encourage musical participation. Prior work indicates that interaction is crucial to robot music systems being considered as true instruments \cite{weinberg2007robotic}, a sentiment shared by our musician consultants.}

\subsection{Design Principles \highlight{for Robot Music Systems}}

\highlight{Addressing \textbf{R3},} we offer five design principles for future design work of robotic percussion musical systems based on \textit{participant} values \highlight{in all (non-interactive and interactive) elements of the robotic performance}.

\subsubsection{\textbf{Create a multi-sensory experience}} Robot artists have an ability to integrate multiple senses in art-making \cite{zhuo2021human}, and it is important to leverage that strength. Our work demonstrates that participants appreciated visual elements, such as \highlight{movement} and lights. Incorporating \highlight{multi-}sensory stimuli can create a more engaging and enjoyable experience.

\subsubsection{\textbf{Incorporate elements of playfulness}} Our findings show that participants responded \highlight{particularly} positively to playful elements, describing the experience as ``cute'' and ``funny''. \highlight{As robots are often stereotyped as cold and mechanical, playful behaviors can create a more engaging, joyful atmosphere.}

\subsubsection{\textbf{Design the system to encourage user interaction}} Interactive elements enhance the \highlight{musical} experience by helping participants feel like active contributors \highlight{and engaging non-experts in music}. However, participants initially hesitated due to unclear interaction opportunities. \highlight{Robot music systems should naturally invite interaction through clear, intuitive cues.}

\subsubsection{\textbf{Emphasize unique strengths of robot performers}} \highlight{Our robots used whole-body movement and} randomness \cite{weinberg2006toward, weinberg2007robotic} \highlight{to perform music in a uniquely robotic way}. Unlike humans, who rely on sensory cues for synchronization \cite{drake2000development}, robots can \highlight{more easily} perform complex rhythms. Participants appreciated \highlight{these diverse capabilities}, \highlight{noting} how robots \highlight{generated rhythmic complexities and coordinated their movements}.

\subsubsection{\textbf{Consider \highlight{the} target \highlight{audience} background}} We observed \highlight{that expertise significantly shaped} how participants perceived the robotic quartet, \highlight{with different elements engaging experts and non-experts in unique ways. System design should thoughtfully incorporate elements that can engage diverse audiences through  different routes to appreciation.}

\subsection{Limitations \& Future Work}

Because we recruited via open call, most participants were likely already interested in robotic music. Additionally, most participants were laypeople in both music and robotics, \highlight{and even ``experts'' were university students studying those fields}. Future research should aim to recruit participants with more diverse opinions and more expertise in these two fields.

\highlight{Additionally, our one-day study could not account for the potential novelty effect, and some survey questions were phrased in a way that may have incurred positively biased responses. Future work should investigate the longer-term impact of robotic music and use more neutral question framing.}

To improve the system, future research should focus on experimenting with different robots or different control methods for more granular control, as suggested by participants. Future investigations could \highlight{also} modify the arena and incorporate a wider range of instruments, including unpitched (e.g., shakers, cymbals) and pitched percussion instruments (e.g., xylophones) to introduce new melodic and rhythmic elements.

Lastly, we propose continued \highlight{discussion and} collaboration with musician\highlight{s} for iterative refinement. This extended partnership could lead to more sophisticated performances that better align with musician values and expectations.

\section{Conclusion}

In this paper, we present the \textit{Beatbots}, a multi-robot percussion quartet co-designed with three musicians. The system uses true percussive rhythms gathered from percussionists, leader-follower dynamics, and randomness to generate unique robotic musical pieces. We held a public performance and surveyed participants ($N=28$). Results indicated that participants \highlight{were receptive to the robot performance, citing common themes of playful engagement, the systems' aesthetic appeal, and appreciation of the music and complex rhythms. Responses} varied by expertise, with expert musicians and non-expert roboticists \highlight{having the greatest mindset shifts. Finally, we discussed affordances inspired by feedback from our musicians for future robot music systems and proposed five design principles for human-interactive robotic music systems.}

\section*{Acknowledgment}
The authors would like to graciously thank Radhika Nagpal for her guidance and support throughout this project.

% and Hyun Kim for assistance with data analysis. We would also like to thank the Princeton School of Engineering and Applied Sciences as well as the Princeton Council on Science and Technology for their financial support of this work.

% Put funding acknowledgment in the first page footnote.

\bibliographystyle{IEEEtrans}
\balance
\bibliography{IEEEabrv,biblio}

\begin{thebibliography}{10}
\providecommand{\url}[1]{#1}
\csname url@rmstyle\endcsname
\providecommand{\newblock}{\relax}
\providecommand{\bibinfo}[2]{#2}
\providecommand\BIBentrySTDinterwordspacing{\spaceskip=0pt\relax}
\providecommand\BIBentryALTinterwordstretchfactor{4}
\providecommand\BIBentryALTinterwordspacing{\spaceskip=\fontdimen2\font plus
\BIBentryALTinterwordstretchfactor\fontdimen3\font minus \fontdimen4\font\relax}
\providecommand\BIBforeignlanguage[2]{{%
\expandafter\ifx\csname l@#1\endcsname\relax
\typeout{** WARNING: IEEEtran.bst: No hyphenation pattern has been}%
\typeout{** loaded for the language `#1'. Using the pattern for}%
\typeout{** the default language instead.}%
\else
\language=\csname l@#1\endcsname
\fi
#2}}

\bibitem{hoffmann2021emotions}
J.~D. Hoffmann, Z.~Ivcevic, and N.~Maliakkal, ``Emotions, creativity, and the arts: Evaluating a course for children,'' \emph{Empirical Studies of the Arts}, vol.~39, no.~2, pp. 123--148, 2021.

\bibitem{gemeinboeck2013creative}
P.~Gemeinboeck and R.~Saunders, ``Creative machine performance: Computational creativity and robotic art.'' in \emph{ICCC}.\hskip 1em plus 0.5em minus 0.4em\relax Citeseer, 2013, pp. 215--219.

\bibitem{jeon2017robotic}
M.~Jeon, ``Robotic arts: Current practices, potentials, and implications,'' \emph{Multimodal Technologies and Interaction}, vol.~1, no.~2, p.~5, 2017.

\bibitem{thorn2020human}
O.~Th{\"o}rn, P.~Knudsen, and A.~Saffiotti, ``Human-robot artistic co-creation: a study in improvised robot dance,'' in \emph{2020 29th IEEE International conference on robot and human interactive communication (RO-MAN)}.\hskip 1em plus 0.5em minus 0.4em\relax IEEE, 2020, pp. 845--850.

\bibitem{weinberg2006toward}
G.~Weinberg and S.~Driscoll, ``Toward robotic musicianship,'' \emph{Computer Music Journal}, pp. 28--45, 2006.

\bibitem{weinberg2007robotic}
G.~Weinberg, ``Robotic musicianship-musical interactions between humans and machines,'' in \emph{Human Robot Interaction}.\hskip 1em plus 0.5em minus 0.4em\relax IntechOpen, 2007.

\bibitem{crick2006synchronization}
C.~Crick, M.~Munz, and B.~Scassellati, ``Synchronization in social tasks: Robotic drumming,'' in \emph{ROMAN 2006-The 15th IEEE international symposium on robot and human interactive communication}.\hskip 1em plus 0.5em minus 0.4em\relax IEEE, 2006, pp. 97--102.

\bibitem{savery2021shimon}
R.~Savery, L.~Zahray, and G.~Weinberg, ``Shimon sings-robotic musicianship finds its voice,'' \emph{Handbook of Artificial Intelligence for Music: Foundations, Advanced Approaches, and Developments for Creativity}, pp. 823--847, 2021.

\bibitem{hoffman2011interactive}
G.~Hoffman and G.~Weinberg, ``Interactive improvisation with a robotic marimba player,'' \emph{Autonomous Robots}, vol.~31, pp. 133--153, 2011.

\bibitem{kapur2007integrating}
A.~Kapur, E.~Singer, M.~S. Benning, G.~Tzanetakis, and Trimpin, ``Integrating hyperinstruments, musical robots \& machine musicianship for north indian classical music,'' in \emph{Proceedings of the 7th international conference on New interfaces for musical expression}, 2007, pp. 238--241.

\bibitem{kusuda2008toyota}
Y.~Kusuda, ``Toyota's violin-playing robot,'' \emph{Industrial Robot: An International Journal}, vol.~35, no.~6, pp. 504--506, 2008.

\bibitem{candy2002integrating}
L.~Candy, E.~Edmonds, and M.~Quantrill, ``Integrating computers as explorers in art practice,'' \emph{Explorations in Art and Technology}, pp. 225--230, 2002.

\bibitem{gomez2021robot}
C.~Gomez~Cubero, M.~Pekarik, V.~Rizzo, and E.~Jochum, ``The robot is present: Creative approaches for artistic expression with robots,'' \emph{Frontiers in Robotics and AI}, vol.~8, p. 662249, 2021.

\bibitem{bruun2020human}
E.~P. Bruun, I.~Ting, S.~Adriaenssens, and S.~Parascho, ``Human--robot collaboration: a fabrication framework for the sequential design and construction of unplanned spatial structures,'' \emph{Digital Creativity}, vol.~31, no.~4, pp. 320--336, 2020.

\bibitem{albin2012musical}
A.~Albin, G.~Weinberg, and M.~Egerstedt, ``Musical abstractions in distributed multi-robot systems,'' in \emph{2012 IEEE/RSJ International Conference on Intelligent Robots and Systems}.\hskip 1em plus 0.5em minus 0.4em\relax IEEE, 2012, pp. 451--458.

\bibitem{rowe2004machine}
R.~Rowe, \emph{Machine musicianship}.\hskip 1em plus 0.5em minus 0.4em\relax MIT press, 2004.

\bibitem{poirson2007integration}
E.~Poirson, J.-F. Petiot, and J.~Gilbert, ``Integration of user perceptions in the design process: application to musical instrument optimization,'' 2007.

\bibitem{clapping}
\BIBentryALTinterwordspacing
J.~Colannino, F.~Gómez, and G.~T. Toussaint, ``Analysis of emergent beat-class sets in steve reich's "clapping music" and the yoruba bell timeline,'' \emph{Perspectives of New Music}, vol.~47, no.~1, pp. 111--134, 2009. [Online]. Available: \url{http://www.jstor.org/stable/25652402}
\BIBentrySTDinterwordspacing

\bibitem{glass}
\BIBentryALTinterwordspacing
J.~Liberatore. Music in all directions: An examination of style in music of philip glass. [Online]. Available: \url{https://performingarts.nd.edu/news-announcements/music-in-all-directions-an-examination-of-style-in-music-of-philip-glass/}
\BIBentrySTDinterwordspacing

\bibitem{ice2012percussion}
S.~P. Ice, \emph{The percussion quartet: A chronological listing and performance guide of six selected works}.\hskip 1em plus 0.5em minus 0.4em\relax The University of Oklahoma, 2012.

\bibitem{camburn2017design}
B.~Camburn, V.~Viswanathan, J.~Linsey, D.~Anderson, D.~Jensen, R.~Crawford, K.~Otto, and K.~Wood, ``Design prototyping methods: state of the art in strategies, techniques, and guidelines,'' \emph{Design Science}, vol.~3, p. e13, 2017.

\bibitem{hartenberger2016cambridge}
R.~Hartenberger, \emph{The Cambridge companion to percussion}.\hskip 1em plus 0.5em minus 0.4em\relax Cambridge University Press, 2016.

\bibitem{benzon1993stages}
W.~L. Benzon, ``Stages in the evolution of music,'' \emph{Journal of Social and Evolutionary Systems}, vol.~16, no.~3, pp. 273--296, 1993.

\bibitem{adler1999mathematics}
C.~Adler, ``Mathematics, automation and intuition in signals intelligence for percussion,'' \emph{System}, vol.~23, no.~2, pp. 19--30, 1999.

\bibitem{langston1989six}
P.~Langston, ``Six techniques for algorithmic music composition,'' in \emph{Proceedings of the International Computer Music Conference}, vol.~60.\hskip 1em plus 0.5em minus 0.4em\relax Citeseer, 1989, p.~59.

\bibitem{mcalpine1999making}
K.~McAlpine, E.~Miranda, and S.~Hoggar, ``Making music with algorithms: A case-study system,'' \emph{Computer Music Journal}, vol.~23, no.~2, pp. 19--30, 1999.

\bibitem{tucker2017emergence}
Z.~Tucker, ``Emergence and complexity in music,'' 2017.

\bibitem{popoff2011indeterminate}
A.~Popoff, ``Indeterminate music and probability spaces: the case of john cage’s number pieces,'' in \emph{Mathematics and Computation in Music: Third International Conference, MCM 2011, Paris, France, June 15-17, 2011. Proceedings 3}.\hskip 1em plus 0.5em minus 0.4em\relax Springer, 2011, pp. 220--229.

\bibitem{reich}
\BIBentryALTinterwordspacing
K.~R. Schwarz, ``Steve reich: Music as a gradual process: Part i,'' \emph{Perspectives of New Music}, vol.~19, no. 1/2, pp. 373--392, 1980. [Online]. Available: \url{http://www.jstor.org/stable/832600}
\BIBentrySTDinterwordspacing

\bibitem{hartenberger2016performance}
R.~Hartenberger, \emph{Performance practice in the music of steve reich}.\hskip 1em plus 0.5em minus 0.4em\relax Cambridge University Press, 2016.

\bibitem{isac2020repetitive}
I.~Isac, ``Repetitive minimalism in the work of philip glass. composition techniques,'' \emph{Bulletin of the Transilvania University of Bra{\c{s}}ov, Series VIII: Performing Arts}, vol.~13, no. 2-Suppl, pp. 141--148, 2020.

\bibitem{feisst2009john}
S.~M. Feisst, G.~Solis, and B.~Nettl, ``John cage and improvisation: an unresolved relationship,'' \emph{Musical improvisation: Art, education, and society}, vol.~2, no.~5, 2009.

\bibitem{cage}
\BIBentryALTinterwordspacing
S.~Z. Solomon. John cage, composed improvisation for snare drum (1987). [Online]. Available: \url{http://szsolomon.com/john-cage-composed-improvisation-snare-drum-1987/}
\BIBentrySTDinterwordspacing

\bibitem{cone2019robots}
E.~Cone and J.~Lambert, ``How robots change the world,'' 2019.

\bibitem{jimeno2019fewer}
J.~F. Jimeno, ``Fewer babies and more robots: economic growth in a new era of demographic and technological changes,'' \emph{SERIEs}, vol.~10, no.~2, pp. 93--114, 2019.

\bibitem{bretan2016survey}
M.~Bretan and G.~Weinberg, ``A survey of robotic musicianship,'' \emph{Communications of the ACM}, vol.~59, no.~5, pp. 100--109, 2016.

\bibitem{kapur2005history}
A.~Kapur, ``A history of robotic musical instruments,'' in \emph{ICMC}, vol.~10, no. 1.88, 2005, p. 4599.

\bibitem{jorda2002afasia}
S.~Jord{\`a}, ``Afasia: the ultimate homeric one-man-multimedia-band,'' in \emph{Proceedings of the 2002 conference on New interfaces for musical expression}, 2002, pp. 1--6.

\bibitem{dannenberg2011mcblare}
R.~B. Dannenberg, H.~B. Brown, and R.~Lupish, ``Mcblare: a robotic bagpipe player,'' \emph{Musical Robots and Interactive Multimodal Systems}, pp. 165--178, 2011.

\bibitem{solis2006waseda}
J.~Solis, K.~Chida, K.~Taniguchi, S.~M. Hashimoto, K.~Suefuji, and A.~Takanishi, ``The waseda flutist robot wf-4rii in comparison with a professional flutist,'' \emph{Computer Music Journal}, pp. 12--27, 2006.

\bibitem{hoffman2010shimon}
G.~Hoffman and G.~Weinberg, ``Shimon: an interactive improvisational robotic marimba player,'' in \emph{CHI'10 Extended Abstracts on Human Factors in Computing Systems}, 2010, pp. 3097--3102.

\bibitem{weinberg2006robot}
G.~Weinberg and S.~Driscoll, ``Robot-human interaction with an anthropomorphic percussionist,'' in \emph{Proceedings of the SIGCHI conference on Human Factors in computing systems}, 2006, pp. 1229--1232.

\bibitem{uchiyama2023development}
J.~Uchiyama, T.~Hashimoto, H.~Ohta, Y.~Nishio, J.-Y. Lin, S.~Cosentino, and A.~Takanishi, ``Development of an anthropomorphic saxophonist robot using a human-like holding method,'' in \emph{2023 IEEE/SICE International Symposium on System Integration (SII)}.\hskip 1em plus 0.5em minus 0.4em\relax IEEE, 2023, pp. 1--6.

\bibitem{zhang2011musical}
A.~Zhang, M.~Malhotra, and Y.~Matsuoka, ``Musical piano performance by the act hand,'' in \emph{2011 IEEE international conference on robotics and automation}.\hskip 1em plus 0.5em minus 0.4em\relax IEEE, 2011, pp. 3536--3541.

\bibitem{wu2010towards}
Y.~Wu, P.~Kuvinichkul, P.~Y. Cheung, and Y.~Demiris, ``Towards anthropomorphic robot thereminist,'' in \emph{2010 IEEE International Conference on Robotics and Biomimetics}.\hskip 1em plus 0.5em minus 0.4em\relax IEEE, 2010, pp. 235--240.

\bibitem{schutz2008seeing}
M.~Schutz, ``Seeing music? what musicians need to know about vision,'' 2008.

\bibitem{pessanha2021virtual}
T.~R.~P. Pessanha, H.~Camporez, J.~Manzolli, B.~S. Masiero, L.~Costalonga, and T.~F. Tavares, ``Virtual robotic musicianship: Challenges and opportunities,'' in \emph{Proceedings of the Sound and Music Computing Conference (SMC’21)}.\hskip 1em plus 0.5em minus 0.4em\relax Sound and Music Computing Network, 2021.

\bibitem{solis2011musical}
J.~Solis and K.~Ng, ``Musical robots and interactive multimodal systems: An introduction,'' in \emph{Musical Robots and Interactive Multimodal Systems}.\hskip 1em plus 0.5em minus 0.4em\relax Springer, 2011, pp. 1--12.

\bibitem{auslander2009lucille}
P.~Auslander, ``Lucille meets guitarbot: Instrumentality, agency, and technology in musical performance,'' \emph{Theatre Journal}, vol.~61, no.~4, pp. 603--616, 2009.

\bibitem{zhuo2021human}
F.~Zhuo, ``Human-machine co-creation on artistic paintings,'' in \emph{2021 IEEE 1st International Conference on Digital Twins and Parallel Intelligence (DTPI)}.\hskip 1em plus 0.5em minus 0.4em\relax IEEE, 2021, pp. 316--319.

\bibitem{flo2015doppelganger}
A.~B. Fl{\o} and H.~Wilmers, ``Doppelg{\"a}nger: A solenoid-based large scale sound installation.'' in \emph{NIME}, 2015, pp. 61--64.

\bibitem{vear2024jess}
C.~Vear, A.~Hazzard, S.~Moroz, and J.~Benerradi, ``Jess+: Ai and robotics with inclusive music-making,'' in \emph{Proceedings of the CHI Conference on Human Factors in Computing Systems}, 2024, pp. 1--17.

\bibitem{kujala2003user}
S.~Kujala, ``User involvement: a review of the benefits and challenges,'' \emph{Behaviour \& information technology}, vol.~22, no.~1, pp. 1--16, 2003.

\bibitem{zamenopoulos2018co}
T.~Zamenopoulos and K.~Alexiou, \emph{Co-design as collaborative research}.\hskip 1em plus 0.5em minus 0.4em\relax Bristol University/AHRC Connected Communities Programme, 2018.

\bibitem{turchet2018smart}
L.~Turchet, ``Smart musical instruments: vision, design principles, and future directions,'' \emph{IEEE Access}, vol.~7, pp. 8944--8963, 2018.

\bibitem{friedman1996value}
B.~Friedman, ``Value-sensitive design,'' \emph{interactions}, vol.~3, no.~6, pp. 16--23, 1996.

\bibitem{sellen2009reflecting}
A.~Sellen, Y.~Rogers, R.~Harper, and T.~Rodden, ``Reflecting human values in the digital age,'' \emph{Communications of the ACM}, vol.~52, no.~3, pp. 58--66, 2009.

\bibitem{halloran2009value}
J.~Halloran, E.~Hornecker, M.~Stringer, E.~Harris, and G.~Fitzpatrick, ``The value of values: Resourcing co-design of ubiquitous computing,'' \emph{CoDesign}, vol.~5, no.~4, pp. 245--273, 2009.

\bibitem{benford2012supporting}
S.~Benford, P.~Tolmie, A.~Y. Ahmed, A.~Crabtree, and T.~Rodden, ``Supporting traditional music-making: designing for situated discretion,'' in \emph{Proceedings of the ACM 2012 conference on Computer Supported Cooperative Work}, 2012, pp. 127--136.

\bibitem{guest_applied_2012}
\BIBentryALTinterwordspacing
G.~Guest, K.~MacQueen, and E.~Namey, \emph{Applied {Thematic} {Analysis}}.\hskip 1em plus 0.5em minus 0.4em\relax 2455 Teller Road, Thousand Oaks California 91320 United States: SAGE Publications, Inc., 2012. [Online]. Available: \url{https://methods.sagepub.com/book/applied-thematic-analysis}
\BIBentrySTDinterwordspacing

\bibitem{mchugh_interrater_2012}
M.~L. McHugh, ``Interrater reliability: the kappa statistic,'' \emph{Biochemia medica}, vol.~22, no.~3, pp. 276--282, 2012.

\bibitem{drake2000development}
C.~Drake, M.~R. Jones, and C.~Baruch, ``The development of rhythmic attending in auditory sequences: attunement, referent period, focal attending,'' \emph{Cognition}, vol.~77, no.~3, pp. 251--288, 2000.

\end{thebibliography}

\end{document}